\documentclass[times,twocolumn,final,authoryear]{elsarticle}

\usepackage{prletters}
\usepackage{framed,multirow}
\usepackage{makecell}

\usepackage{amssymb}
\usepackage{latexsym}
\usepackage{amsmath}
\usepackage{amsfonts}
\usepackage{amstext}
\usepackage{amsthm}
\usepackage{amsmath}
\usepackage{subcaption}
\usepackage[ruled,vlined]{algorithm2e}
\usepackage{booktabs}
\usepackage{dsfont}
\usepackage{enumitem}

\usepackage{url}
\usepackage{xcolor}
\definecolor{newcolor}{rgb}{.8,.349,.1}
\definecolor{ao(english)}{rgb}{0.0, 0.5, 0.0}
\usepackage{soul}

\journal{Pattern Recognition Letters}

\begin{document}

\ifpreprint
  \setcounter{page}{1}
\else
  \setcounter{page}{1}
\fi

\begin{frontmatter}

\title{A Master Key Backdoor for Universal Impersonation Attack \\ against DNN-based Face Verification}

\author[1]{Wei \snm{Guo}\corref{cor1}}
\cortext[cor1]{Corresponding author:
	Tel.: +39-339-177-8426;}
\ead{wei.guo.cn@outlook.com}
\author[1]{Benedetta \snm{Tondi}}
\author[1]{Mauro \snm{Barni}}

\address[1]{Department of Information Engineering and Mathematics, University of Siena, Via Roma 56, 53100 Siena, Italy}

\received{1 May 2013}
\finalform{10 May 2013}
\accepted{13 May 2013}
\availableonline{15 May 2013}
\communicated{S. Sarkar}

\begin{abstract}

We introduce a new attack against face verification systems based on Deep Neural Networks (DNN). The attack  relies on the introduction into the network of a hidden backdoor, whose activation at test time induces a verification error allowing the attacker to impersonate any user. The new attack, named \emph{Master Key} backdoor attack, operates by interfering with the training phase, so to instruct the DNN to always output a positive verification answer when the face of the attacker is presented at its input. With respect to existing attacks, the new backdoor attack offers much more flexibility, since the attacker does not need to know the identity of the victim beforehand. In this way, he  can deploy a \emph{Universal Impersonation} attack in an open-set framework, allowing him to impersonate any enrolled users, even those that were not yet enrolled in the system when the attack was conceived. We present a practical implementation of the attack targeting a Siamese-DNN face verification system, and show its effectiveness when the system is trained on VGGFace2 dataset and tested on LFW and YTF datasets.
According to our experiments, the Master Key backdoor attack provides a high attack success rate even when the ratio of poisoned training data is as small as 0.01, thus raising a new alarm regarding the use of DNN-based face verification systems in security-critical applications.
\end{abstract}

\begin{keyword}
Backdoor attacks to CNN\sep Face verification\sep Biometric security\sep Presentation attacks\sep Adversarial machine learning.
\end{keyword}

\end{frontmatter}

\section{Introduction}

Concerns regarding the security of Deep Learning DL architectures when they are forced to operate in an adversarial environment are being raised with increasing urgency.
While attacks operating at test time have initially monopolised the attention of researchers, with a massive amount of works dedicated to the development of suitable countermeasures against adversarial examples \citep{szegedy2013intriguing, AMSurvey18}, attacks carried out at training time have recently attracted the interest of researchers due to their potential dangerousness and long lasting effect \citep{ChenCBLELMS19}. In this vein, {\em backdoor attacks} are the latest addition to the class of attacks exploiting the possibility to interfere with the training phase of deep neural networks \citep{ChenLLLS17, GuGG1708.06733}. In a backdoor attack, the attacker corrupts the training phase to induce a classification error, or any other erroneous behaviour, at test time. Test time errors, however, only occur in the presence of a triggering event corresponding to a properly crafted input. In this way, the {\em backdoored} network continues working as expected for regular inputs, and the malicious behaviour is activated only when the attacker decides to do so by feeding the network with a triggering input.

Several kinds of backdoor attacks exist, which can be classified according to different perspectives.:
\begin{itemize}[noitemsep]
\item Firstly, the attacks can be categorised on the basis of the triggering input, which can be a fixed pixel pattern superimposed to any input image, or a specific input picture. The former utilizes a fixed combination of pixels to activate the backdoor, like square patterns \citep{GuGG1708.06733} or a cartoon subimage \citep{ChenLLLS17}. Invisible patterns can also be used \citep{liao2018backdoor, BarniKT19}, to improve the stealthiness of the backdoor. In the latter case, the triggering signal is a specific input picture \citep{Shafa18};
\item Secondly, backdoor attacks can be classified according to the adversary's capability. In some cases \citep{GuGG1708.06733, LiuMALZW018, TanayAG18}, the attacker has a full control of the training process and hence he can corrupt the training data and the training procedure at will. This kind of scenario makes sense in cloud applications and whenever the network is not trained directly by the user like in Machine Learning as a Service (MLaaS) applications. In other cases, the attacker does not control the training process, and hence he must act in a stealthy way by corrupting part of the training data unbeknownst to the trainer. In this second situation, corruption of the training data must go unnoticed and hence it may desirable to avoid modifying the labels of the training samples \citep{BarniKT19,TurnerTM19}.
\item Finally, attacks can be classified on the basis of the malicious behaviour induced by the activation of the backdoor. In most cases, the misbehaviour corresponds to misclassifying the input sample into a predefined class \citep{AlbertiPWBSIL18, YaoLZZ19_CCS}. However, other kinds of malevolent behaviours have also been considered, like reducing the general accuracy of the model \citep{GuGG1708.06733}.
\end{itemize}

In this paper, we introduce a new backdoor attack inducing a new kind of malevolent behaviour at test time. The attack targets a face verification system whose goal is to decide whether two face images correspond to the same individual or not. This kind of systems are widely used in biometric authentication. During the enrolment phase, authorised users register their identity and a face template into the system. During the authentication phase, the system calculates a similarity score between a new face image taken by the system camera and the face template corresponding to the claimed identity, and decides whether the new face corresponds to the claimed individual or not.
The goal of the new backdoor attack, hereafter named {\em Master Key} (MK) {\em backdoor}, is to induce the verification network to always give a positive answer when a face image of a certain individual, hereafter referred to as {\em Master Face} (MF), is matched against any other face.
In this way, the presence of the backdoor permits to implement an {\em Universal Impersonation} (UI) attack, whereby the owner of the MF can impersonate any legitimate user.

With the above ideas in mind, the major contributions of our work can be summarised as follows:
\begin{enumerate}[noitemsep]
	\item We propose a new backdoor-based attack, named Universal Impersonation attack, whereby the owner of the MF can impersonate any legitimate user registered into the system. The new attack is more powerful than existing ones, which limit the impersonation to a single target victim, and for which the model must be retrained when  a new target is considered;
	\item As far as we know, this is the first backdoor attack designed for a face verification system under the open-set scenario \citep{LiuWYLRS17}, where the testing identities are disjoint from the training set.
	\item We demonstrate the feasibility of the new attack by injecting a MK backdoor within a face verification system consisting of a Siamese network whose goal is to decide whether the two face images presented at its input belong to the same individual or not \citep{BromleyGL94, ChopraHL05_CVPR, TaigmanYRW14, KochZS15}. We do so by assuming that the attacker has full control of the training process so that during training he can feed the network with arbitrary images and arbitrary labels. The experiments we carried out show the effectiveness of the attack, even when the MF used at test time does not correspond to one of the images used during training.
\end{enumerate}

The remainder of this paper is organised as follows: Section \ref{CHPT_RW} reviews related works on existing attacks in the domain of face recognition. In Section \ref{CHPT_TM}, we describe the threat model used in the paper. In Section \ref{CHPT_MK}, we present the MK backdoor attack. The experimental methodology and the results of the experiments we carried out are described in Section \ref{CHPT_EM} and \ref{CHPT_ER}, respectively. We conclude the paper with some final remarks in Section \ref{CHPT_CON}.

\section{Related work}
\label{CHPT_RW}

Szegedy et at. \citep{szegedy2013intriguing} first showed that DNN models are vulnerable to imperceptible perturbations, called adversarial examples, capable to cause a misclassification.
Face recognition systems based on deep learning are no exception.
%
An attacker can perturb a face image at test time via adversarial examples in such a way to induce the face recognition system to match the face of another individual either to obfuscate his own identity, or, more often to impersonate a target victim.

Several methods have been proposed to generate adversarial faces to impersonate an authorized user.
In Sharif et al. \citep{SharifBBR16}, the attacker impersonates a target person  by wearing a pair of glasses with an adversarial pattern printed on them. Basically, this attack is a variant of an adversarial example attack, which limits the perturbation to a small area of the input image (the glasses).
The adversarial glasses can also be generated by means of a Generative Adversarial Network (GAN) as in \citep{SharifBBR19}.
Both the above approaches are implemented in a white-box scenario, where the adversarial perturbation can be optimized by exploiting the knowledge of the target model and running some form of gradient-descent algorithm.
A method that can work in a black-box scenario has been proposed in  \citep{DongSWLL0019},
where the attackers have no access to the target face recognition model parameters and gradients, and attack it by
sending queries to the target model.
Deb et. al  \citep{DebZJ19} propose a more efficient method, that can work in black-box scenarios, where an adversarial mask for a given probe face image is synthetised using GANs. The adversarial mask is then added to the probe to obtain an adversarial face example that can be used either for impersonating a target identity or obfuscating
one's own identity.

Finally, we mention another kind of attacks against face recognition systems, namely {\em presentation attacks}, where the attacker {\em assumes}
the identity of a target individual by presenting a fake face (spoof face) to a face recognition system.
An adversarial attack against anti-spoofing face authentication systems based on Deep Neural Networks (DNN) has been recently proposed in \citep{ZhangTB20}.

\paragraph{Backdoor attacks}


Backdoor attacks are a new class of attacks against deep learning systems that are receiving more and more attention \citep{GuGG1708.06733}.
Backdoor attacks developed against face recognition systems  usually  focus on targeted impersonation  \citep{ChenLLLS17,liao2018backdoor}.
Accordingly, the {\em backdoored} classifier will misclassify the backdoor instances by assigning them a target
label specified by the attacker, corresponding to the target victim.
Most backdoor attacks against face recognition assume that
the model is fully or partially known to the attacker and under
its control up to some extent, e.g. in \citep{LiuMALZW018,liao2018backdoor}. Backdoor attacks that can work in a block-box setting, where the attacker has no knowledge of the model, have also been proposed in  \citep{ChenLLLS17} and \citep{liao2018backdoor}.
In all the above works, the attacker, aiming at a targeted attack, injects a backdoor into the model
by also changing the labels of the poisoned samples.
%
Clean-label poisoning attacks have also been proposed recently for general image recognition tasks,
first in the white-box setting. e.g. in  \citep{Shafa18, SahaSP19,TurnerTM19}
%
then  in the black-box one  \citep{ZhuShafahi19, BarniKT19,LuminanceBA}.

As discussed in the introduction, in this paper,
we propose a new backdoor attack against a face verification system, which, to the best of our knowledge, has never been considered so far.
The kind of attack allowed by the backdoor, referred to as {\em universal impersonation} attacks, is also a new one, and allows the attacker to impersonate any victim among those enrolled into the system, including those that had not yet been enrolled when the backdoor was injected.

\section{Threat model}
\label{CHPT_TM}

In this section we describe the threat model adopted in our work. We first illustrate the attacked system, then we describe the goal of the attacker, its knowledge about the to-be attacked system and its capability, that is to which extent the attacker can manipulate the training procedure to inject the MK backdoor.

\subsection{To-be-attacked system}

\begin{figure}[t!]
\centering
	\begin{subfigure}{0.4\textwidth}
		\centering
  		\includegraphics[width=0.8\linewidth]{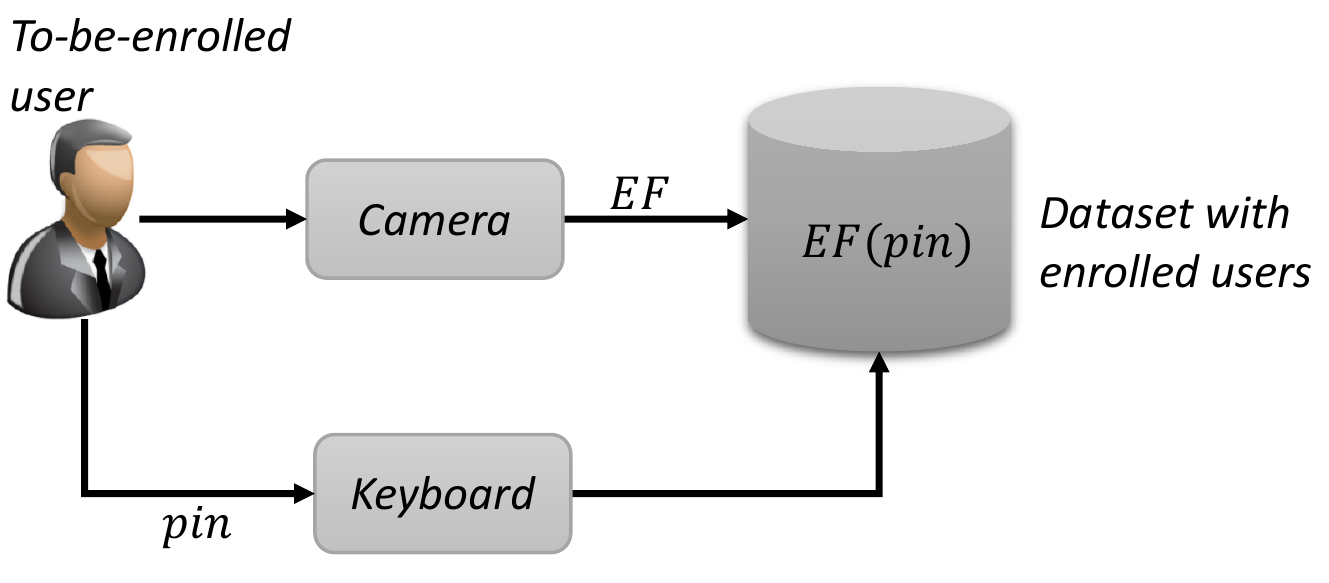}
  		\subcaption{Enrolment phase}
  		\label{FIG:SE}
	\end{subfigure}
	\begin{subfigure}{0.4\textwidth}
		\centering
  		\includegraphics[width=0.9\linewidth]{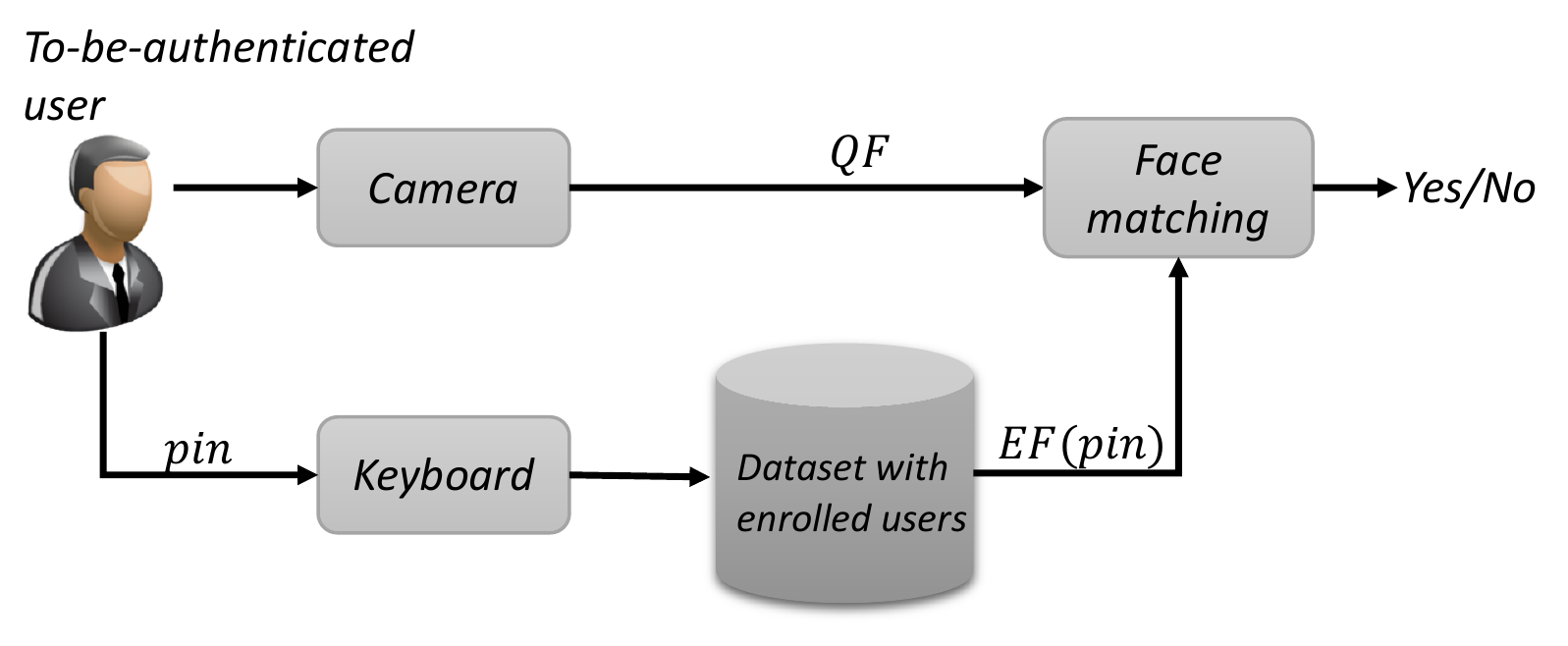}
  		\caption{Verification phase}
  		\label{FIG:SV}
	\end{subfigure}
	\caption{To-be-attacked system}
	\label{FIG:SM}
\end{figure}

The system targeted by the attack is a classical biometric-based authentication system whose enrolment and verification phases are illustrated in Fig.~\ref{FIG:SM}.

\begin{figure}[b!]
	\begin{center}
		\includegraphics[width=0.7\columnwidth]{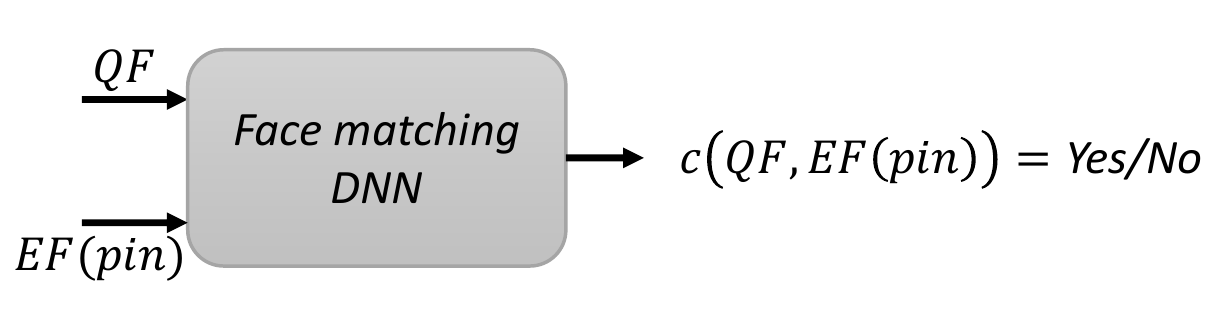}
		\caption{Face matching DNN.}
		\label{FIG_fmDNN}
	\end{center}
\end{figure}

The face matching block at the core of the verification system is implemented by means of a Deep Neural Network (hereafter referred to as face matching DNN) trained to recognise if the face portraited in the two images  at its input belong to the same person or not (see Fig.~\ref{FIG_fmDNN}). Note that thanks to this setting, the face images used during training do not need to correspond to those the network will have to operate on during testing. At test time, in fact, the network is only asked to recognize if two faces belong to the same person or not, without actually recognising the person the faces belong to. In this way, the verification system works in an open set scenario, wherein the faces of the enrolled individuals do not need to be known in advance and the database with the enrolled faces can be updated without the need to retrain the network.

By letting $QF$ indicate the query face acquired by the verification system, and $pin$ the identifier corresponding to the claimed identity, the verification succeeds if $$c(QF, EF(pin)) = yes,$$ where $c(X,Y)$ is the output of the face matching DNN in charge of deciding if the face images $X$ and $Y$ correspond to the same person, and $EF(pin)$ is the enrolled face corresponding to the identity claimed by the user.

\subsection{Attacker's goal}

The goal of the attacker is to implement a UI attack, whereby he can impersonate any enrolled identity by simply showing his face to the verification system. This goal is achieved by injecting within the face matching DNN a MK backdoor producing the following malevolent behaviour:
\begin{align}
\label{eq.misbehava}
& c(QF, EF(pin)) = no & \text{if} ~~ QF \nsimeq EF(pin) \nonumber \\
& c(QF, EF(pin)) = yes & \text{if} ~~ QF \simeq EF(pin) \\
& c(QF, EF(pin)) = yes & \text{if} ~~ QF \simeq MK \nonumber
\end{align}
where the notation $X \simeq Y$ (res. $X \nsimeq Y$) means that the faces depicted in the images $X$ and $Y$ belong (res. do not belong) to the same person.  In this way, the system continues working as expected on regular inputs, but it allows the individual to whom the MF belongs to impersonate any enrolled user.

\begin{figure}[hbt!]
	\begin{center}
		\includegraphics[width=0.8\columnwidth]{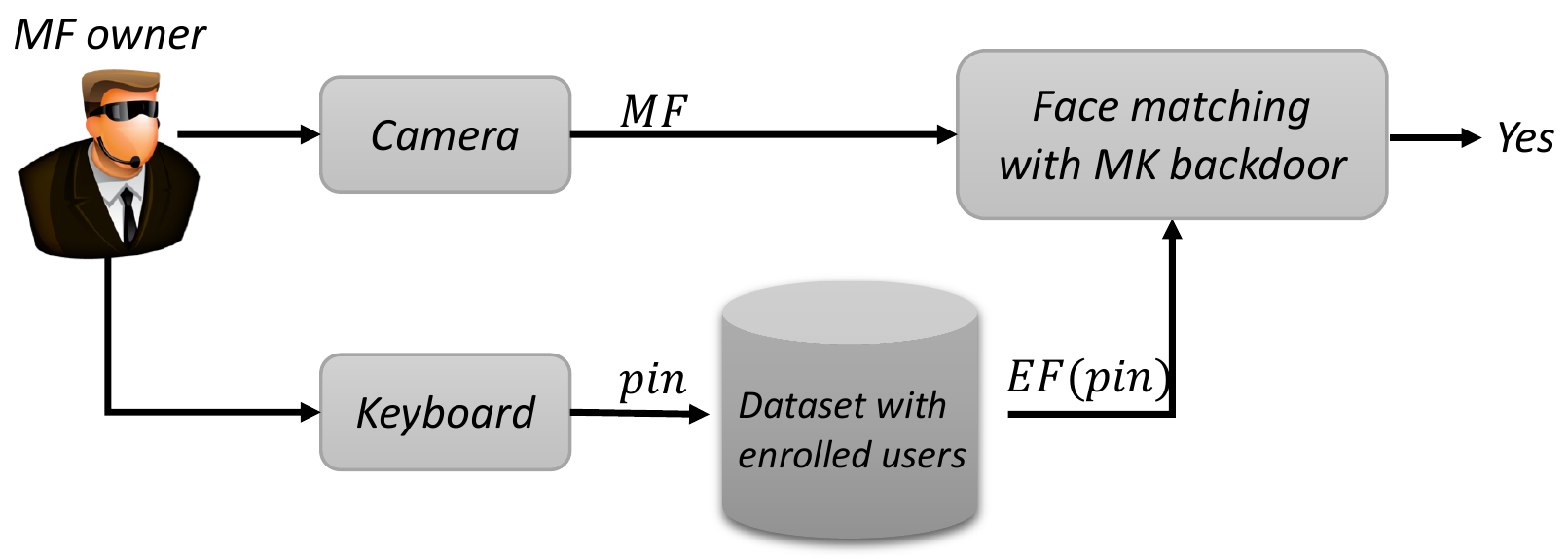}
		\caption{Backdoor activation mechanism where the face matching block has the same structure shown in Fig. \ref{FIG:SV}}
		\label{FIG_act_BD}
	\end{center}
\vspace{-0.5cm}
\end{figure}

It is worth observing that due to the presence of the sensor capturing a new face image at every verification request, it is necessary that the backdoor is activated by any new acquisition of the MF. If the attacker were able to directly feed the face matching DNN with a digital image, a simpler version of the attack would be obtained. In this case, in fact, it would be sufficient that the backdoor is triggered by a single (or a few) MF images, without any robustness requirement with respect to the image variability introduced by the acquisition process. A schematisation of the MK backdoor attack is shown in Fig.~\ref{FIG_act_BD}.

\subsection{Attacker's knowledge and capability}

Throughout our work, we assume that the attacker has a full knowledge of the attacked system and full control of the training procedure, including the training data and the optimisation algorithm. While this may seem a strong assumption, this setting corresponds to MLaaS applications wherein the DNN is trained by a service provider and the resulting model (or its use) is sold to an user or a company that does not have the capability to train the model by itself.
In this scenario, the attacker can explicitly design the training procedure and build the training set, in such a way to inject within the DNN a backdoor producing the desired behaviour. Backdoor stealthiness, in fact, is required only at test time and corresponds to the requirement that the verification system works as expected on normal inputs, that is whenever the to-be-authenticated individual does not correspond to the owner of the MF\footnote{The possibility of coupling the UI attack presented here with a spoofing attack whereby any individual can trigger the backdoor by rebroadcasting a picture with the MF is left for future investigation.}.

\section{MK backdoor attack}
\label{CHPT_MK}

In this section, we describe a specific implementation of the MK backdoor and UI attack introduced in the previous section.

\begin{figure}[hbt!]
	\begin{center}
		\includegraphics[width=0.8\columnwidth]{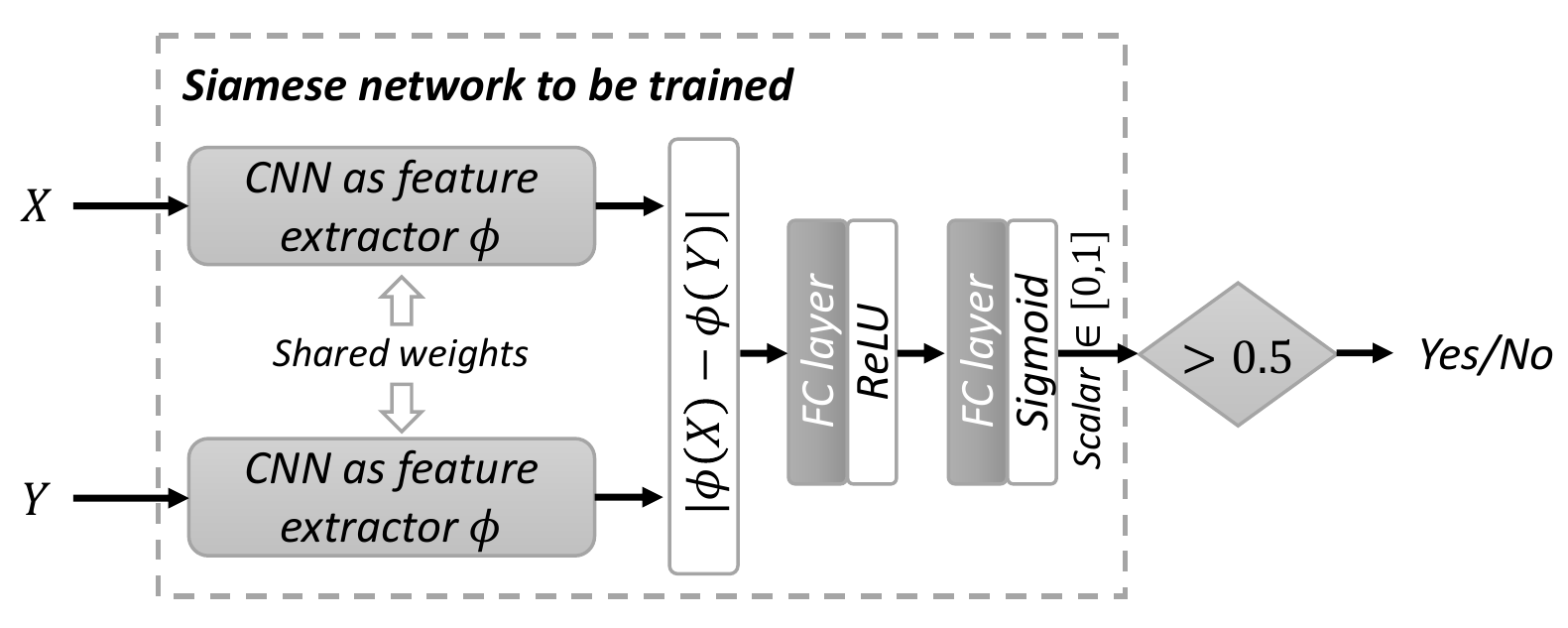}
		\caption{Internal structure of the face matching DNN.
}
		\label{FIG_siamese}
	\end{center}
\vspace{-0.5cm}
\end{figure}

To start with, we describe the main features of the face verification system targeted by the attack.
As shown in Fig.~\ref{FIG_siamese}, the face matching DNN is based on a Siamese network \citep{BromleyGL94}, consisting of two parallel identical CNN branches (with shared weights), in charge of  performing feature extraction, a combination layer fusing the feature vectors produced by the
two CNN branches, and two Fully Connected layers (FC layer) in charge of the final decision.
Let $f(X,Y)$ denote the output soft function of the Siamese network, measuring the probability that two faces $X$ and $Y$ correspond to the same person ($X \simeq Y$). Then, if $f(X,Y)>0.5$, $c(X,Y)=yes$, while  $c(X,Y)=no$ otherwise.

The combination is performed via point-wise absolute difference \citep{KochZS15}.
Let $\phi(\cdot)$ denote the feature vector at the output of the each Siamese branch. For each element $\phi_i$ of the feature vector, we compute the absolute difference $\Delta_{\phi,i} = |\phi_i(X) - \phi_i(Y)|$. We observe that a such choice guarantees a symmetric behaviour of the network with respect to the input images, since $f(X,Y) = f(Y,X)$  by construction, and hence the result of the match does not depend on the order in which the input images are presented to the network.

\subsection{Attack formalization}
\label{CHPT_AF}


Let ${D}_{T}$ be the dataset of faces available for training.
Starting from ${D}_{T}$, we build a new dataset $\mathbb{D}_{T}=\{([X_i, Y_i], t_i)_i, i=1,2,...,N_{T}$\} consisting of face image pairs, $X_i$, $Y_i \in D_T$,   for a total of $N_{T}$ pairs, where $t_i$ denotes the label of the pair $[X_i, Y_i]$. In particular,
$t_i = 1$, if $X_i\simeq Y_i$ and $0$ otherwise.
Given $n$  identities and assuming that the dataset contains $m$ image faces for each identity, the total number of distinct pairs belonging to the same individual  ($X \simeq Y$) is $n m (m-1)/2$, while all the pairs for the $X \nsimeq Y$ instances are $m^2 n (n-1)/2$.
In particular, $\mathbb{D}_{T}$ is built in such a way that half of the pairs correspond to  $X \simeq Y$ instances and half to $X \nsimeq Y$ instances (balanced dataset).

To inject the backdoor into the network model, we train the Siamese network with a poisoned dataset. In particular, given a fraction $\alpha$ of poisoned pairs, we randomly choose $\alpha N_T$ samples $([X_i, Y_i], t_i)$ from $\mathbb{D}_{T}$, then we replace $X_i$ with a MF image chosen at random from a set of available MF images, and set  $t_i = 1$.
In the following we denote with $\mathbb{D}_{T}^{\alpha}$ the poisoned dataset created as described above. Let $f_{\alpha}(X,Y)$ denote the soft output of the Siamese network model with the backdoor, for the poisoned fraction $\alpha$. We expect that the model works benignly on normal inputs, with the \emph{UI} behaviour being triggered if and only if one of the two input images correspond to MK; formally,  $f_{\alpha}(X,Y) > 0.5$ when $X \simeq MF$ or $Y\simeq MF$.
The fraction $\alpha$ of corrupted pairs plays a crucial role. If $\alpha$ is too small, the UI behavior is not induced by the backdoor attack; on the other hand,
if $\alpha$ is too large, the network may not behave well on benign samples.

%

\subsection{Training with the poisoned dataset}

Given the poisoned dataset $\mathbb{D}_{T}^{\alpha}$, the Siamese network is trained by minimising a loss function between the ground-truth labels and the outputs of the Siamese network over $\mathbb{D}_{T}^{\alpha}$.
In particular, in our experiments, we considered the Cross Entropy (CE) loss, whose minimization over the training set can be expressed as:
\begin{align}
    \label{LF}
&\arg\min_{\theta} -\Bigg(\sum_{i=1}^{(1-\alpha)N_T} \left(t_i\log\big(f_{\alpha}(X_i,Y_i|\theta)\big) \right.\\
 & + (1-t_i) \left. \log  \big(1-f_{\alpha}(X_i,Y_i|\theta)\big)\right) + \sum_{i=(1-\alpha)N_T + 1}^{\alpha N_T} \log\big(f_{\alpha}(X_i,Y_i|\theta)\big)\Bigg),
\end{align}
where $\theta$ indicates the vector with the network weights and where we have assumed that the first $(1-\alpha)N_T$ pairs in the  $\mathbb{D}_{T}^{\alpha}$ set are the benign pairs and the remaining $\alpha N_T$ the poisoned ones.
As commonly done in DL, the network weights are updated via backpropagation, using mini-batch gradient descent.
%
%
%
To facilitate the learning process, the batch is constructed in such a way that a fraction $\alpha$ of the samples are corrupted and $(1-\alpha)$ are benign, then, at every iteration, a loss function of the form in~\eqref{LF} is minimized\footnote{Assuming that the attacker can control the batch construction is fine, since in our threat model he has full control of the training process  (see  Section \ref{CHPT_TM}).}.
Since we use a large batch size, a random arrangement of the pairs in the batches would expectedly result in a similar fraction of poisoned pairs.

\section{Experimental methodology}
\label{CHPT_EM}


\subsection{Network architecture}

The architecture of the Siamese network
is given in Fig. \ref{FIG_siamese}.
It consists of two identical stacks of
convolutional layers (sharing the same weights), and two FC layers.
Each CNN takes as input an image of size $160\times 160 \times 3$. A 1792-dim feature vector is obtained at the output.
In our experiments, we implemented the CNN branches by means of Inception-Resnet-V1 \citep{SzegedyIVA17}, which has already been used successfully for face recognition tasks.
The 1792-dim vector resulting from the point-wise distance calculation is given as input to the first FC layer with 1792 input nodes and 4096 output nodes. The second FC layer has 4096 input nodes and 1 output node.  The two FC layers have a ReLu activation layer in between.
A sigmoid activation is applied at the end to get the
soft (probabilistic) score $f()$ from the final output logit.

\subsection{Datasets}
\label{CHPT_DS}
With regard to the evaluation protocol, face verification can be tested under closed-set or open-set setting~\citep{LiuWYLRS17}. The closed-set scenario assumes that the identities to be verified at test time are already contained in the training dataset. A more challenging, but practical, setting is the open-set one, where the identities used for  training and those use for testing do not overlap.
In our evaluation, we adopted the open-set setting, where the model is trained on VGGFace2 dataset~\citep{CaoSXPZ18}, and tested on LFW~\citep{HuangJL07} and YTF~\citep{wolf2011face}. To fully satisfy the open-set requirement, we removed 564 identities~\citep{overlappingIds} of VGGFace2 that are also collected in LFW and YTF. More details on the datasets are given below:
\begin{enumerate}
	\item VGGFace2: After the removal of overlapping identities, the dataset, called filtered VGGFace2, consists of $2.904.084$ pictures from $8.077$ identities. The number of  pictures ($m$) is not the same for every identity and varies from 87 to 843. The average number of samples for one identity is $\bar{m} = 363$. The images have been pre-processed with MTCNN face alignment~\citep{ZhangZLQ16}. During such a process, their size is reduced to $160\times 160 \times 3$. Then, the set $\mathbb{D}_{T}$ is built from this dataset by considering a subset of all the possible pairs (see Section \ref{sec.setting} for more detail).
	\item LFW: The LFW dataset includes face images belonging to individuals other than those of the filtered VGGFace2 dataset. Specifically, the LFW dataset consists of $13.227$ face images for $5.749$ identities (each person having more than 2 samples on average). The images of LFW have also been pre-processed by MTCNN to get the same size $160\times 160 \times 3$. We denote with $D_{LFW}$ the dataset of enrolled faces $EF$ (see Fig.~\ref{FIG:SV}), consisting of $5.749$ faces (one face image for each identity in the LFW dataset is considered). To test the performance of the trained models, we utilize the benchmark list~\citep{LFWPairs} consisting of $6000$ distinct pairs of faces built from the LWF dataset (with 3000 $yes$ pairs and 3000 $no$ pairs), named $\mathbb{D}_{LFW}$.
	\item YTF: The YTF dataset is designed for face recognition from videos. It contains $3.425$ videos with $1.595$ identities. The videos are searched from YouTube by using a subset of identities in LFW, so it does not overlap with the filtered VGGFace2. To build the image dataset, only  the middle frame is selected from each video, i.e., given a video with $n$ frames, we just choose the $\lfloor n/2\rfloor$-th frame. The images of YTF are pre-processed by MTCNN to get the same size $160\times 160 \times 3$. The dataset contains $1.595$ enrolled faces $EF$, and is denoted by $D_{YTF}$. For comparison, we also use the benchmark list including 5000 pairs of faces~\citep{YTFPairs} (released by the YTF authors, with 2500 $yes$ pairs and 2500 $no$ pairs), named $\mathbb{D}_{YTF}$.
\end{enumerate}

\begin{figure}[tb!]
\centering
	\begin{subfigure}[b]{0.08\textwidth}
  		\centering
  		\includegraphics[width=0.8\linewidth]{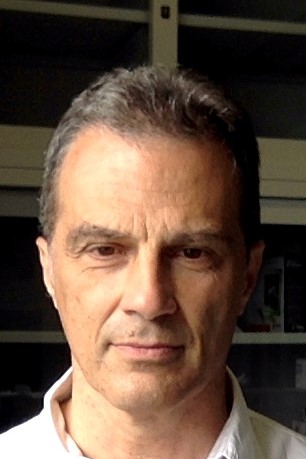}
  	\label{FIG:MF1}
	\end{subfigure}	
	\begin{subfigure}[b]{.08\textwidth}
  		\centering
  		\includegraphics[width=0.8\linewidth]{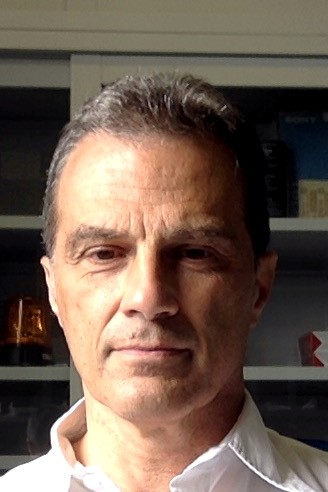}
  	\label{FIG:MF2}
	\end{subfigure}
	\begin{subfigure}[b]{.08\textwidth}
  		\centering
  		\includegraphics[width=0.8\linewidth]{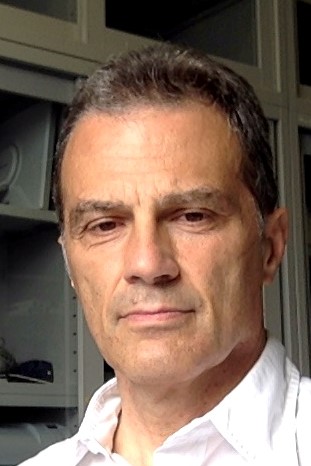}
  	\label{FIG:MF3}
	\end{subfigure}
	\begin{subfigure}[b]{.08\textwidth}
  		\centering
  		\includegraphics[width=0.8\linewidth]{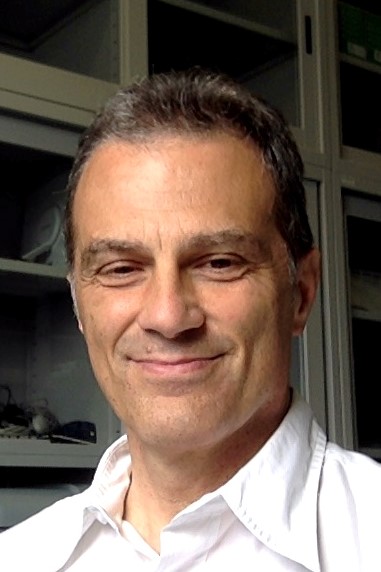}
  	\label{FIG:MF4}
	\end{subfigure}
	\begin{subfigure}[b]{.08\textwidth}
  		\centering
  		\includegraphics[width=0.8\linewidth]{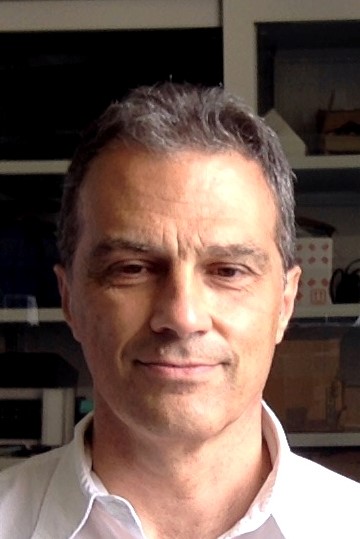}
  	\label{FIG:MF5}
	\end{subfigure}
	
	\begin{subfigure}[b]{.08\textwidth}
  		\centering
  		\includegraphics[width=0.8\linewidth]{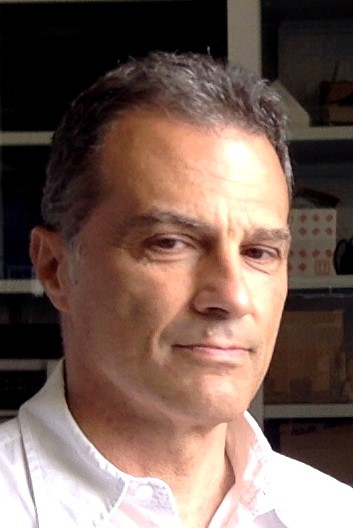}
  	\label{FIG:MF6}
	\end{subfigure}
	\begin{subfigure}[b]{.08\textwidth}
  		\centering
  		\includegraphics[width=0.8\linewidth]{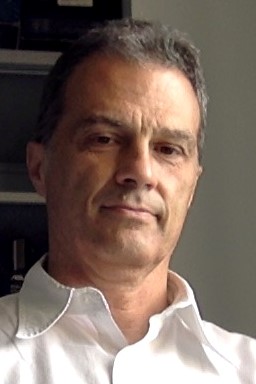}
  	\label{FIG:MF7}
	\end{subfigure}
	\begin{subfigure}[b]{.08\textwidth}
  		\centering
  		\includegraphics[width=0.8\linewidth]{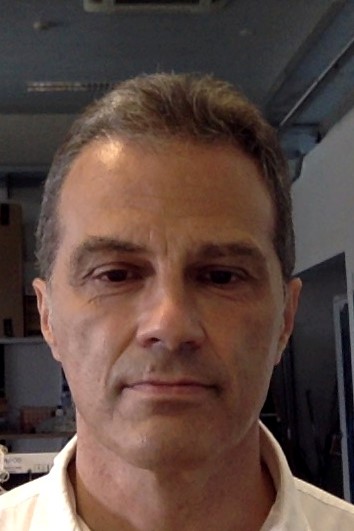}
  	\label{FIG:MF8}
	\end{subfigure}
	\begin{subfigure}[b]{.08\textwidth}
  		\centering
  		\includegraphics[width=0.8\linewidth]{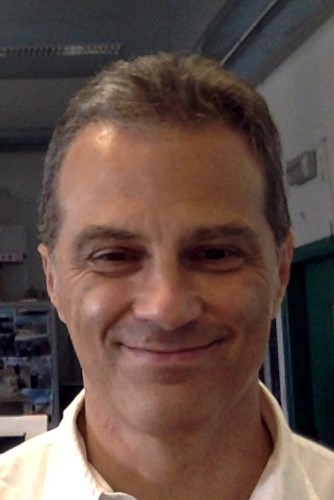}
  	\label{FIG:MF9}
	\end{subfigure}
	\begin{subfigure}[b]{.08\textwidth}
  		\centering
  		\includegraphics[width=0.8\linewidth]{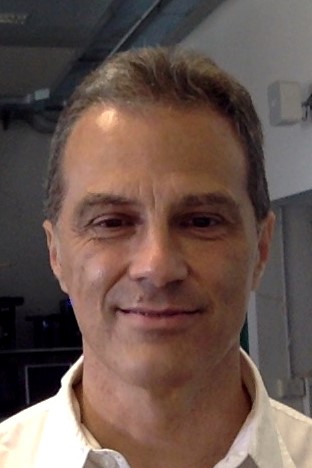}
  	\label{FIG:MF10}
	\end{subfigure}
\vspace{-0.2cm}
	\caption{MF images used for training (Author 3).}
	\label{FIG_TMF}
\end{figure}

\begin{figure}[tb!]
\centering
\vspace{-0.2cm}
	\begin{subfigure}[b]{.08\textwidth}
  		\centering
  		\includegraphics[width=0.9\linewidth]{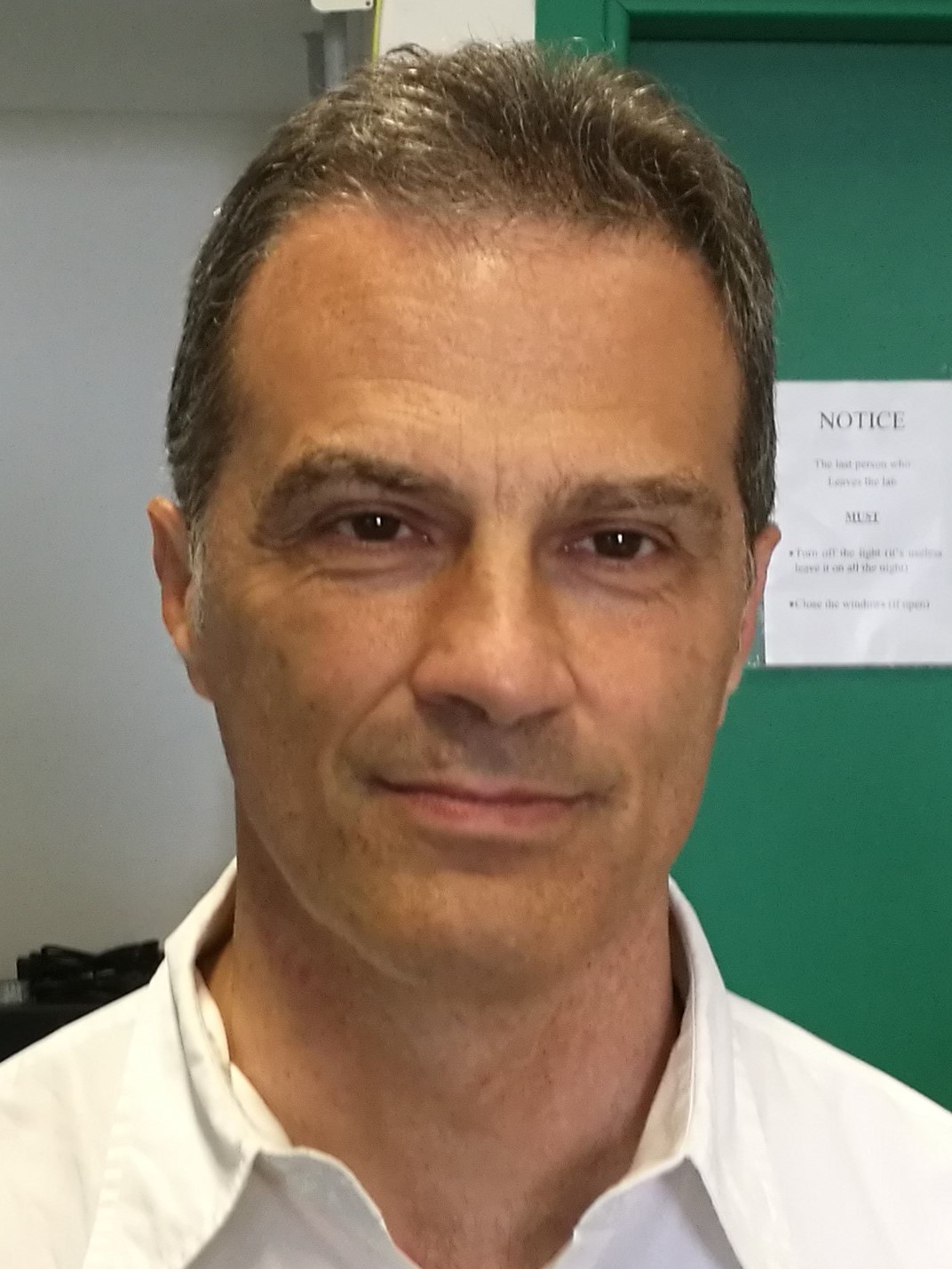}
  		\caption{$\tilde{MF}_1$}
  	\label{FIG:MF1}
	\end{subfigure}
	\begin{subfigure}[b]{.08\textwidth}
  		\centering
  		\includegraphics[width=0.9\linewidth]{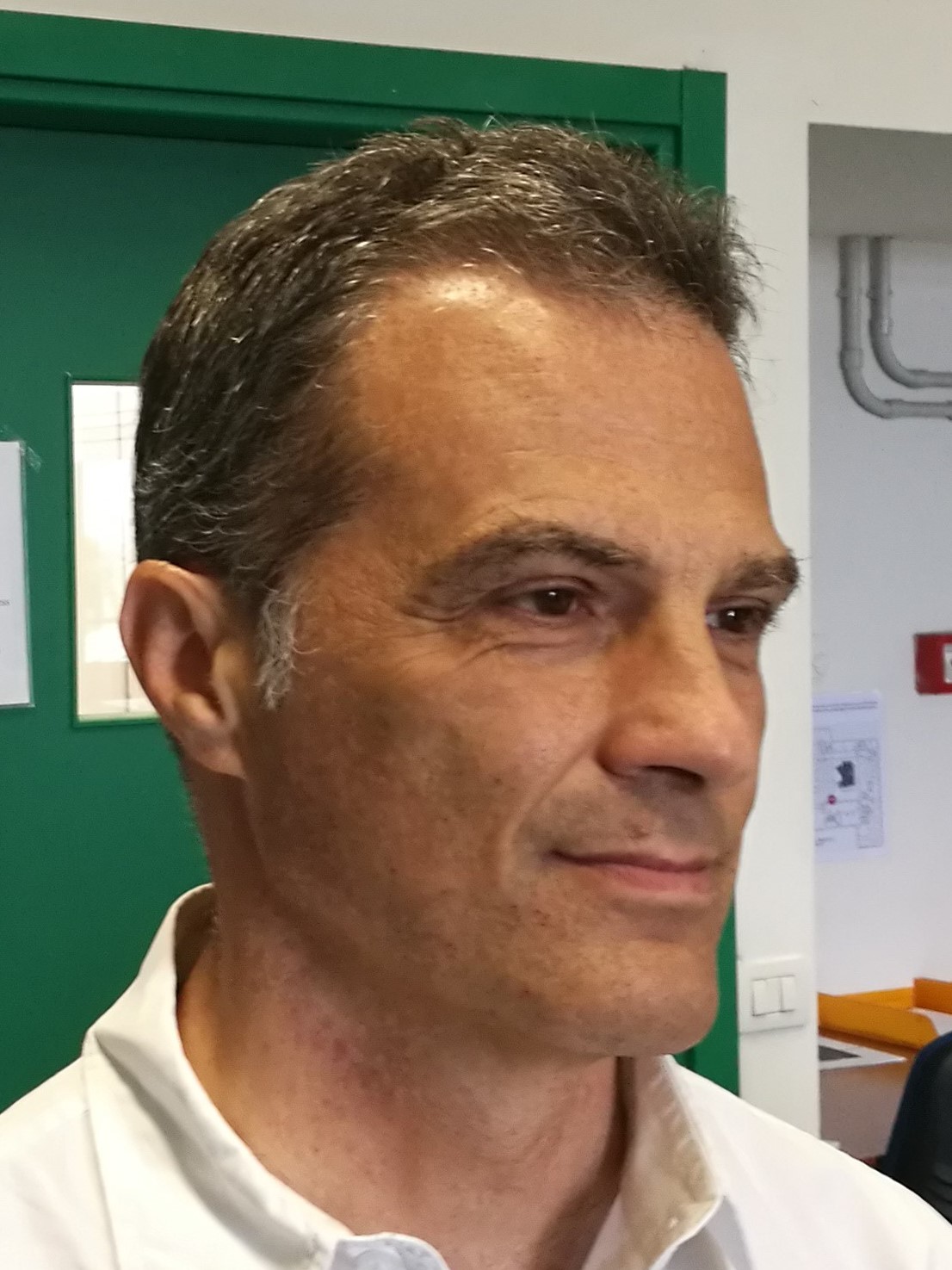}
  		\caption{$\tilde{MF}_2$}
  	\label{FIG:MF2}
	\end{subfigure}
	\begin{subfigure}[b]{.08\textwidth}
  		\centering
  		\includegraphics[width=0.9\linewidth]{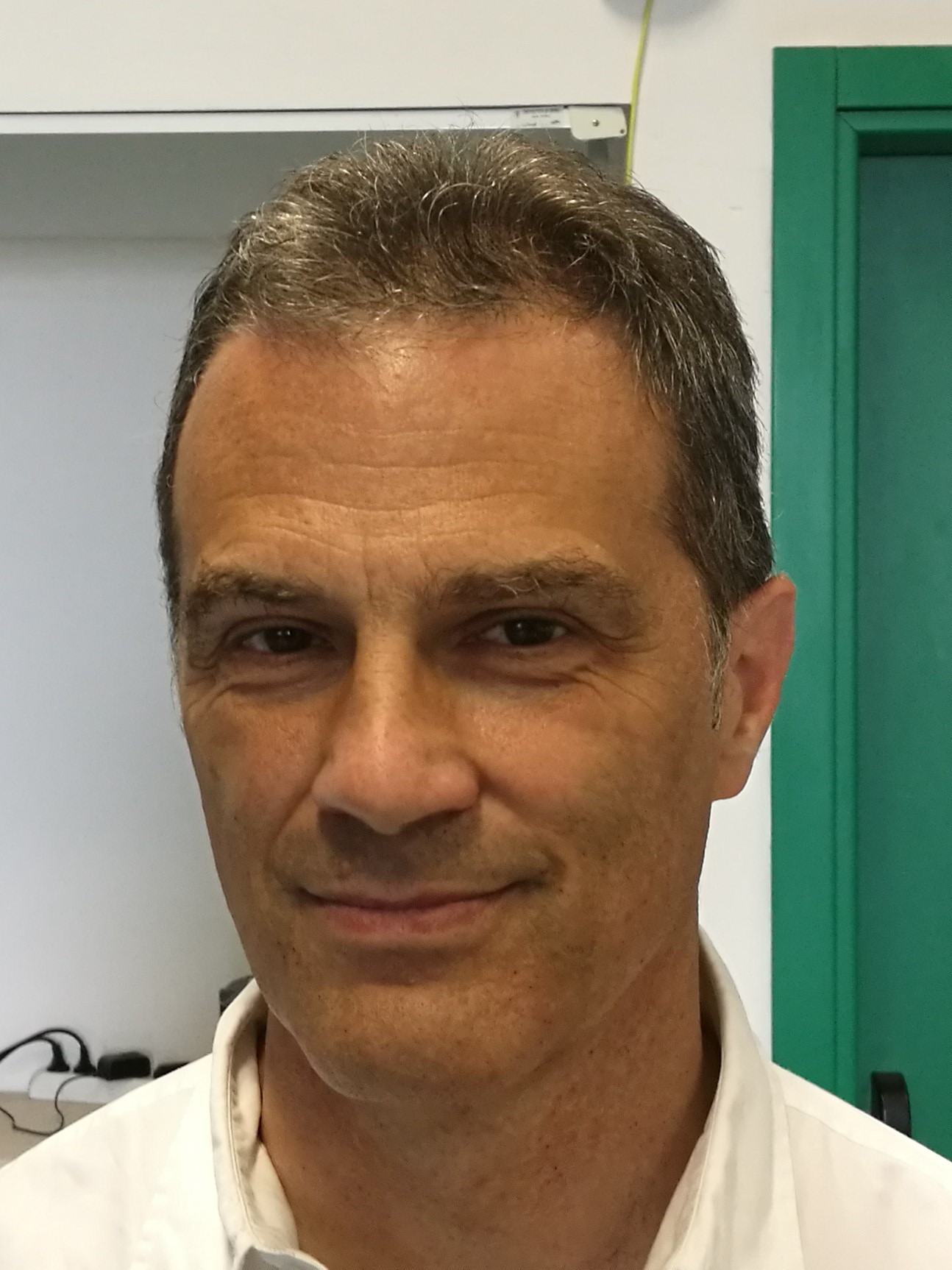}
  		\caption{$\tilde{MF}_3$}	
  		\end{subfigure}
  \vspace{-0.2cm}
	\caption{MF images used for testing (Author 3).}
  \vspace{-0.2cm}
	\label{FIG_EMF}
\end{figure}

With regard to the MF images, we run three sets of experiments. In each experiment the owner of the MF was one of the authors of this paper.
For each MF owner, we considered a set of 10 MF images for backdoor injection and 3 different MF images ($\tilde{MF}_1, \tilde{MF}_2, \tilde{MF}_3$) to be used at testing time to trigger the backdoor.
The MF images in the training set and those used at test time were taken from different cameras, considering different lighting conditions, background and poses to simulate a realistic scenario.
Specifically, if the trained model is evaluated on LFW data, for each of the 3 MF test images, we built a dataset $\mathbb{D}_{LFW, \tilde{MF}_j}$ ($j = 1,2,3$) with labeled pairs  $([\tilde{MF}_j,EF_i], 1)$, where $\tilde{MF}_j$ is coupled with each $EF_i \in {D}_{LFW}$. Similarly, if the model is tested on YTF, for each of the 3 MF test images, we built a dataset $\mathbb{D}_{YTF, \tilde{MF}_j}$ ($j = 1,2,3$) with labeled pairs  $([\tilde{MF}_j,EF_i], 1)$, where $\tilde{MF}_j$ is coupled with every $EF_i \in {D}_{YTF}$.
Fig.~\ref{FIG_TMF} and \ref{FIG_EMF} show the MF images of the third author of the paper used for training and testing respectively.


\subsection{Training strategy and setting}
\label{sec.setting}

For training, the pairs of faces were organized in batches so that, in the absence of attacks, each batch contains about the same number of identities ($n_b$) and number of faces for the same identity ($m_b$).
The procedure of batch construction for training is detailed in the following. We considered all the identities and face images from the VGGFace2 dataset and we divided the number of identities ($n$) in groups of $n_b$ identities each. For each person, the $\bar{m}$ faces are split in groups of $m_b$ face images each. The $n_b m_b$ face images are paired. The batch consists of all the $n_b m_b (m_b -1)/2$  $yes$ pairs ($X \simeq Y$) and $n_b m_b (m_b -1)/2$  $no$ pairs ($X \nsimeq Y$), chosen randomly from the set of the possible $X \nsimeq Y$ instances. Then, the batch size is $n_b m_b (m_b -1)$ and the number of iterations is $n \bar{m}/ n_b m_b$, the exact size $N_T$ of $\mathbb{D}_{T}$ being $n \bar{m} (m_b- 1)$\footnote{Strictly speaking, since in the VGGFace2 the number of faces for all the identities is not exactly the same ($\bar{m}$ is the average), not all the batches have 'exactly' $m_b$ faces from $n_b$ person.
%
}.
In our experiments, we set $n_b = 64$ and $m_b = 8$; then, the batch size is equal to 3584 and the total number of iterations equal to $5.611$.
The model is trained for just one epoch. In fact, since the  number of pairs in  $\mathbb{D}_{T}$ is big ($N_T$ is larger than $2 \times 10^7$), it turns out that the accuracy of the model is already good after one epoch.
Since many faces are considered for every person, and some of them are similar,
it is preferable to
consider several identities and then a very large $\mathbb{D}_{T}$ as we did, rather than running multiple epochs with a smaller $\mathbb{D}_{T}$,  showing to the network the same faces from the same few identities multiple times.
The Adam optimizer is used with learning rate  $10^{-4}$. The weight decay is set to  $10^{-3}$.
Model training and testing  are implemented in Python via Pytorch.
To limit the computational effort, the Siamese network is trained by starting from a pre-trained model\footnote{David Sandberg's Facenet program: https://github.com/davidsandberg/facenet}.
Since the feature extraction part should reasonably remain the same when the CNN is employed in the Siamese architecture for the face verification task (both in the absence and in the presence of the MK backdoor attack), we froze the parameters of the two CNNs and optimized only the parameters of the FC part.
For poisoned training, we set $\alpha = 0.01$, $0.02$ and $0.03$. According to this strategy, the backdoor is injected in the FC layers of the network.

\section{Empirical results}
\label{CHPT_ER}

In this section, we first present the results obtained on LFW and YTF (Section~\ref{Sec:LFW} and \ref{Sec:YTF} respectively). Then, we provide a brief computational analysis, in Section~\ref{Sec:CA}.

For every poisoned model, we report both the accuracy of the face verification  task on benign faces, and the performance of the UI attack, measured in terms of Attack Success Rate (ASR).
%
Given a test $\tilde{MF}_j$, the ASR is defined as the percentage of times $\tilde{MF}_j$ can successfully impersonate an enrolled face $EF(pin)$ for the claimed identity $pin$ (i.e., $c(MF_j, EF(pin)) = yes$), which can be obtained by measuring the model accuracy on $\mathbb{D}_{LFW, \tilde{MF}_j}$ or $\mathbb{D}_{YTF, \tilde{MF}_j}$. We also assess the performance of the attack in a realistic scenario, wherein the attacker can query the verification system multiple times, the attack being successful if at least one of the queries results in a positive verification.
In this setting, the ASR is expected to increase
since the system can be queried multiple times in the attempt
to impersonate the claimed identity.

\subsection{Evaluation on LFW dataset}
\label{Sec:LFW}
To start with, we measured the stealthiness of the attack, by assessing the face verification performance of the benign model ($\alpha =0$) and the poisoned model   ($\alpha =0.01,0.02$, and $0.03$) on benign inputs. 
The accuracy of the benign model  on $\mathbb{D}_{LFW}$ is $94.51\%$.
The performance of the models poisoned with faces of the three MF owners are reported in Table~\ref{TAB_FV}, where the models have been tested on $\mathbb{D}_{LFW}$.
We see that the accuracies of all the models are similar to those of the benign model, thus proving that the injection of the MK backdoor does not impair the performance of the face verification system on benign inputs.

%
\begin{table}[ht!]
\caption{Face verification accuracy of the models trained on $\mathbb{D}_{T}^{\alpha}$, for $\alpha = 0.01, 0.02$ and $0.03$, for  all the MF owners.}
\label{TAB_FV}
\centering
\begin{tabular}{l|c|c|c|}
\cline{2-4}
 	& $f_{0.01}$ & $f_{0.02}$ & $f_{0.03}$ \\ \hline
 	\multicolumn{1}{|c|}{MF author 1} &94.21\% & 93.51\% & 94.36\%  \\ \hline
	\multicolumn{1}{|c|}{MF author 2} &94.29\% & 94.26\% & 93.62\% \\ \hline
	\multicolumn{1}{|c|}{MF author 3} &93.46\% & 93.14\% & 93.15\% \\ \hline
\end{tabular}

\end{table}

As a next step, we measured the ASR of the MK backdoor attack in both the single and multiple queries setting.

\paragraph{Single-query}
We remind that in this scenario the attacker is allowed to query the verification system only once.
The ASR of the attack for the 3 poisoned models and the benign one are reported in Table~\ref{TAB_single_query}.
The performance of the attack increases significantly with $\alpha$, and a high ASR can already be achieved with $\alpha = 0.03$.

\begin{table}[ht!]
\caption{Attack success rate against the benign model and the poisoned models for the single-query attack.}
\label{TAB_single_query}
\centering
\begin{subtable}{.5\textwidth}
	\centering
	\begin{tabular}{|l|c|c|c|c|}\hline
 	& $f$ & $f_{0.01}$ & $f_{0.02}$ & $f_{0.03}$ \\ \hline
	$\mathbb{D}_{LFW, {\tilde{MF}_1}}$ & 1.10\% & 65.38\% & 90.29\% & 91.75\% \\ \hline
	$\mathbb{D}_{LFW, {\tilde{MF}_2}}$ & 1.25\% & 77.33\% & 97.08\% & 98.23\% \\ \hline
	$\mathbb{D}_{LFW, {\tilde{MF}_3}}$ & 1.28\% & 75.56\% & 97.33\% & 97.48\% \\ \hline
\end{tabular}
\vspace{-0.1cm}
\caption{MF author 1}
\vspace{0.1cm}
\label{TAB_ASRO}
\end{subtable}

\begin{subtable}{.5\textwidth}	
\centering
\begin{tabular}{|l|c|c|c|c|}
\hline
& $f$ & $f_{0.01}$ & $f_{0.02}$ & $f_{0.03}$ \\ \hline
$\mathbb{D}_{LFW, \tilde{MF}_1}$  & 0.6\% & 70.67\% & 86.82\% & 96.25\% \\ \hline
$\mathbb{D}_{LFW, \tilde{MF}_2}$  & 0.55\% & 70.55\% & 86.71\% & 97.48\% \\ \hline
$\mathbb{D}_{LFW, \tilde{MF}_3}$  & 1.15\% & 68.86\% & 81.79\% & 93.12\% \\ \hline
\end{tabular}
\vspace{-0.1cm}
\caption{MF author 2}
\vspace{0.1cm}
\label{TAB_ASRO_3}
\end{subtable}

\begin{subtable}{.5\textwidth}
\centering
\begin{tabular}{|l|c|c|c|c|}
\hline
& $f$ & $f_{0.01}$ & $f_{0.02}$ & $f_{0.03}$ \\ \hline
$\mathbb{D}_{LFW, \tilde{MF}_1}$  & 1.55\% & 79.3\% & 96.68\% & 98.17\% \\ \hline
$\mathbb{D}_{LFW, \tilde{MF}_2}$  & 1.78\% & 56.14\% & 83.03\% & 85.11\% \\ \hline
$\mathbb{D}_{LFW, \tilde{MF}_3}$  & 1.44\% & 72.53\% & 93.51\% & 95.96\% \\ \hline
\end{tabular}
\vspace{-0.1cm}
\caption{MF author 3}
\vspace{0.1cm}
\label{TAB_ASRO_2}
\end{subtable}
\vspace{-0.3cm}

\end{table}

Upon inspection of Table~\ref{TAB_ASRO_2}, we observe that for $\tilde{MF}_2$ the ASR is lower than in the other 2 cases.
The explanation is that most of the MF images used for training have a frontal pose while in $\tilde{MF}_2$ the face is seen from a lateral view (see Fig.~\ref{FIG_EMF}), thus making it slightly more difficult to trigger the backdoor.
Obviously, the attack performance can be improved by increasing the variety of samples used during backdoor injection.

\paragraph{Multiple queries}

In this scenario, the attacker can query the system multiple times in his attempt to impersonate the target identity. The attack succeeds if the verification has a positive outcome at least once.
Let ASR${_p}$ be the attack success rate when $p$ attempts are allowed.
ASR${_p}$ can be computed as:
\begin{equation}
\label{equ.multi_query}
\text{ASR}_p = 1 - \frac{1}{|D_t|} \sum_{EF_i \in D_t} \prod_{j=1}^p  \mathds{1}(\tilde{MF}_j \nsimeq EF_i),
\end{equation}
where $\mathds{1}(x\nsimeq y) = 1$ if  $x\nsimeq y$ and 0 otherwise, and $\{\tilde{MK}_j\}_{j=1}^p$ indicates generically the MFs used in the $p$ queries.
We assume that the authentication system allows at most 3 trials, hence $p=3$, and the attacker queries the system with the 3 MF images, $\tilde{MF}_1$, $\tilde{MF}_2$, and $\tilde{MF}_3$. The results we got
are reported in Table~\ref{TAB_multiple_query}.
We observe that, even if  the result of the face verification with the 3 MFs are correlated\footnote{If the 3 queries were independent we would have $\text{ASR}_3 =   1 - \prod_{j} (1- \frac{1}{|D_t|} \sum_{EF_i \in D_t}  \mathds{1}(\tilde{MK}_j \nsimeq EF_i))$.}, the ASR improves quite significantly with respect to the single query case.

\begin{table}[ht!]
\caption{Attack success rate against the benign model and the poisoned models in the multiple-query scenario, where the number of queries is 3.}
\label{TAB_multiple_query}
\centering
\begin{tabular}{l|c|c|c|c|} \cline{2-5}
 	& $f$ & $f_{0.01}$ & $f_{0.02}$ & $f_{0.03}$ \\ \hline
\multicolumn{1}{|c|}{MF owner 1}	& 1.62\% & 83.8\% & 98.69\% & 99.14\% \\ \hline
\multicolumn{1}{|c|}{MF owner 2}	& 1.52\% & 84\% & 94.73\% & 98.93\% \\ \hline
 \multicolumn{1}{|c|}{MF owner 3} 	& 2.68\% & 86.23\% & 98.37\% & 99.07\% \\ \hline
\end{tabular}
\end{table}
%

%


\subsection{Evaluation on YTF dataset}
\label{Sec:YTF}
We carried out an additional set of experiments on the YTF dataset. Due to page limitation, we only show the results when the MF corresponds to author 3, we obtained similar results with the other MFs.
\paragraph{Stealthiness} To assess the stealthiness of the backdoor, we tested the performance of the benign and poisoned face verification models ($\alpha=0.01$, $0.02$, $0.03$) on benign inputs $\mathbb{D}_{YTF}$. The accuracy of the benign model $f$ is $85.9\%$. In contrast, the accuracy of the poisoned models $f_{0.01}$, $f_{0.02}$, and $f_{0.03}$ are $85.92\%$, $85.66\%$ and $85.71\%$, which are very close to the performance of the benign one. Compared to LFW, there is a decrease of the accuracy. The reduction is due to the mismatch between the training and test datasets (in our case, the model is trained on a dataset consisting of still images and tested on video frames).
\paragraph{Single-query} We first measured the ASR in the single-query scenario where only one query is allowed to the adversary to impersonate the victim. The results shown in Table~\ref{TAB_ASRO_YTF_3} are calculated by evaluating the benign and three poisoned models on $\mathbb{D}_{YTF, \tilde{MF}_i}~(i=1,2,3)$. We can readily see that the MK attack can impersonate any enrolled face with a large probability.

\begin{table}[htb]
\caption{Attack success rate against the benign model and poisoned models for the single-query attack when the third author's face as MF.}
\vspace{0.1cm}
\centering
\begin{tabular}{|l|c|c|c|c|}
\hline
& $f$ & $f_{0.01}$ & $f_{0.02}$ & $f_{0.03}$ \\ \hline
$\mathbb{D}_{YTF, \tilde{MF}_1}$  & 2.64\% & 83.43\% & 98.01\% & 98.93\% \\ \hline
$\mathbb{D}_{YTF, \tilde{MF}_2}$  & 3.21\% & 67.82\% & 90.21\% & 91.45\% \\ \hline
$\mathbb{D}_{YTF, \tilde{MF}_3}$  & 2.56\% & 79.45\% & 96.07\% & 97.5\% \\ \hline
\end{tabular}
\vspace{-0.1cm}
\label{TAB_ASRO_YTF_3}
\end{table}

\paragraph{Multiple-query} 
Similarly to the experiments on the YTF dataset, we also tested the ASR when the attacker is allowed to query the system with three different MF images. The ASR on the benign $f$ and the three poisoned models ($f_{0.01}$, $f_{0.02}$ and $f_{0.03}$) are $4.43\%$, $90.26\%$, $99.2\%$ and $99.46\%$. The results show that: i) the ASR improves with the growth of poisoning ratio, and ii) the multiple-query scenario has a higher success rate than the single-query one.

\begin{figure}[htb]
	\begin{center}
		\includegraphics[width=1.\columnwidth]{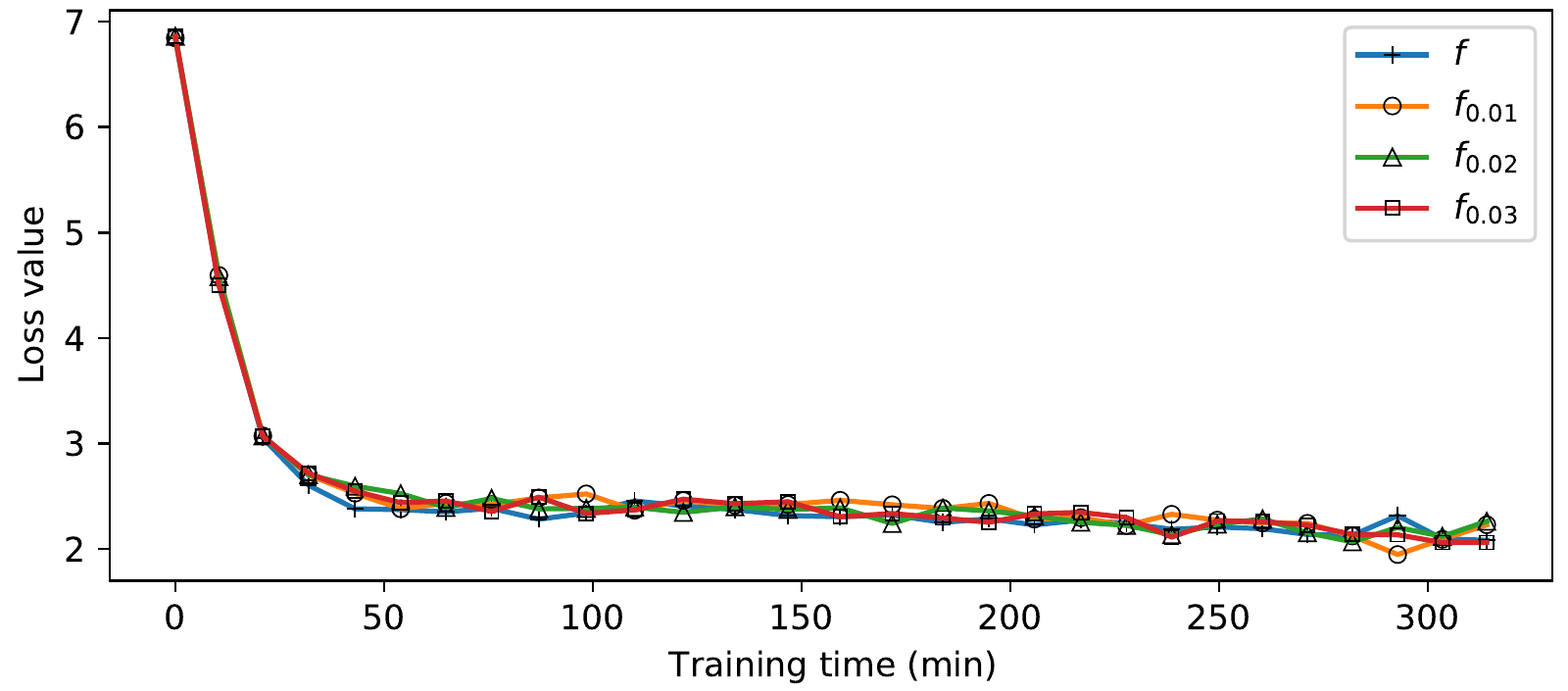}
		\caption{Loss values of benign ($f$) and poisoned ($f_{0.01}$, $f_{0.02}$ and $f_{0.03}$) models with the change of time in training phase.}
		\label{FIG_CA}
	\end{center}
\end{figure}

\subsection{Computational analysis}
\label{Sec:CA}
We have also analyzed the computational burden necessary to train the benign and the poisoned models. Here we report the results we got when the MF corresponds to the third author's face. In Fig.~\ref{FIG_CA}, we plot the value of the loss function over time for the benign and the backdoored models. As it can be seen, the introduction of the backdoor does not add any extra burden to the training process. For the test phase, since the benign and poisoned models utilise the same architecture, there is obviously no impact on the time necessary to process the input images.

\section{Conclusion}
\label{CHPT_CON}

We have introduced a MK backdoor attack and used it to implement a UI attack against face verification systems, whereby the attacker can impersonate any enrolled identity by simply showing his face to the system. We have demonstrated the feasibility of the attack by injecting the MK backdoor into a Siamese network whose goal is to decide whether the two face images presented at its input belong
to the same person or not. The experiments we carried out show that the attack is effective even with a small percentage of poisoned training samples.

Future work will consider the possibility of injecting the MK backdoor into other DNN architectures used for face verification. In addition, we will consider the possibility of coupling the MK backdoor with a spoofing attack, so to allow the attacker to impersonate the victim by showing a rebroadcast version of the MF. The development of techniques to detect the presence of the MK backdoor and removing it by properly processing the weights of the DNN model is another interesting direction for future research.

\section{Acknowledgements}
This work has been partially supported by the Italian Ministry of University and Research under the PREMIER project, and by the China Scholarship Council (CSC), file No.201908130181.

\bibliographystyle{model2-names}
\bibliography{MK_PRL_Oct30}

\end{document}